\documentclass[10pt,twocolumn,letterpaper]{article}

\usepackage{times}
\usepackage{epsfig}
\usepackage{graphicx}
\usepackage{amsmath}
\usepackage{amssymb}

\usepackage{algpseudocode}
\usepackage{algorithm}
\usepackage{subfigure}

\usepackage{color}

\date{}

\begin{document}

\newcommand{\etal}{\textit{et al}.}

\algnewcommand\algorithmicswitch{\textbf{switch}}
\algnewcommand\algorithmiccase{\textbf{case}}
\algnewcommand\algorithmicassert{\texttt{assert}}
\algnewcommand\Assert[1]{\State \algorithmicassert(#1)}%
\algdef{SE}[SWITCH]{Switch}{EndSwitch}[1]{\algorithmicswitch\ #1\ \algorithmicdo}   {\algorithmicend\ \algorithmicswitch}%
\algdef{SE}[CASE]{Case}{EndCase}[1]{\algorithmiccase\ #1}{\algorithmicend\     \algorithmiccase}%
\algtext*{EndSwitch}%
\algtext*{EndCase}%
\algnewcommand\algorithmicinput{\textbf{input:}}
\algnewcommand\Input{\item[\algorithmicinput]}

\title{End-to-end Global to Local CNN Learning for Hand Pose Recovery in Depth Data\footnote{This work has been submitted to the IEEE for possible publication. Copyright may be transferred without notice, after which this version may no longer be accessible.}}

\author{\parbox{16cm}{\centering
    {\large Meysam Madadi$^{1,2}$, Sergio Escalera$^{1,3}$, Xavier Bar\'o$^{1,4}$, Jordi Gonz\`alez$^{1,2}$}\\
    {\normalsize
    $^1$ Computer Vision Center, Edifici O, Campus UAB, 08193 Bellaterra (Barcelona), Catalonia Spain\\
    $^2$ Dept. of Computer Science, Univ. Aut\`onoma de Barcelona (UAB), 08193 Bellaterra, Catalonia Spain\\
    $^3$ Dept. Mathematics and Informatics, Universitat de Barcelona, Catalonia, Spain\\
    $^4$ Universitat Oberta de Catalunya, Catalonia, Spain}}
}

\maketitle

\begin{abstract}
Despite recent advances in 3D pose estimation of human hands, especially thanks to the advent of CNNs and depth cameras, this task is still far from being solved. This is mainly due to the highly non-linear dynamics of fingers, which make hand model training a challenging task. In this paper, we exploit a novel hierarchical tree-like structured CNN, in which branches are trained to become specialized in predefined subsets of hand joints, called local poses. We further fuse local pose features, extracted from hierarchical CNN branches, to learn higher order dependencies among joints in the final pose by end-to-end training. Lastly, the loss function used is also defined to incorporate appearance and physical constraints about doable hand motion and deformation. Finally, we introduce a non-rigid data augmentation approach to increase the amount of training depth data. Experimental results suggest that feeding a tree-shaped CNN, specialized in local poses, into a fusion network for modeling joints correlations and dependencies, helps to increase the precision of final estimations, outperforming state-of-the-art results on NYU and SyntheticHand datasets.
\end{abstract}

\section{Introduction}
\label{sec:intro}
Recently, hand pose recovery attracted special attention thanks to the availability of low cost depth cameras, like Microsoft Kinect \cite{choi2015,Supancic2015,ge2016,qian2014,Oberweger2015iccv,Tan2016,sridhar2013,keskin2011,tang2014,Sun2015}. Unsurprisingly, 3D hand pose estimation plays an important role in most HCI application scenarios, like social robotics and virtual immersive environments \cite{rautaray2011interaction}.

Despite impressive pose estimation improvements thanks to the use of CNNs and depth cameras, 3D hand pose recovery still faces some challenges before becoming fully operational in uncontrolled environments with fast hand/fingers motion, self occlusions, noise, and low resolution \cite{Usabiaga2009}. Although the use of CNNs and depth cameras has allowed to model highly non-linear hand pose motion and finger deformation under extreme variations in appearance and viewpoints, accurate 3D-based hand pose recovery is still an open problem.

Two main strategies have been proposed in the literature for addressing the aforementioned challenges: Model-based and data-driven approaches. Model-based generative approaches fit a predefined 3D hand model to the depth image \cite{Tan2016,qian2014,Sharp2015,makris2015hierarchical,Oikonomidis2011,Gorce2011}. However, as a many-to-one problem, accurate initialization is critical; besides, the use of global objective functions might not convey accurate results in case of self-occlusions of fingers. 

Alternatively, the so-called data-driven approaches consider the available training data to directly learn hand pose from appearance. Data-driven approaches for hand pose estimation have been benefited from recent advances on Convolutional neural networks (CNNs) \cite{xu2013,Supancic2015,Sun2015,tang2014,keskin2011,Kirac2014}. CNNs, as in many other computer vision tasks, have been successfully applied in data-driven hand-pose recovery approaches either for heat-map regression of discrete outputs (corresponding to joint estimation probabilities), or direct regression of continuous outputs (corresponding to joint locations)  \cite{Oberweger2015,Oberweger2015iccv,tompson2014,ge2016,sinha2016deephand}. Heat-map regression models require additional optimization time for computing the likelihood of a joint being located at a particular spatial region. Unfortunately, heat-map based methods are subject to propagate errors when mapping images to final joint space. A main issue with CNNs as direct regression models, on the other hand, is how to deal with high nonlinear output spaces, since too complex models jeopardize generalization. Indeed for CNNs, learning suitable features (i.e. with good generalization and discrimination properties) in highly nonlinear spaces while taking into account structure and dependencies among parameters, is still a challenging task.

\begin{figure}[!t]
  \centering
  \includegraphics[width=\linewidth]{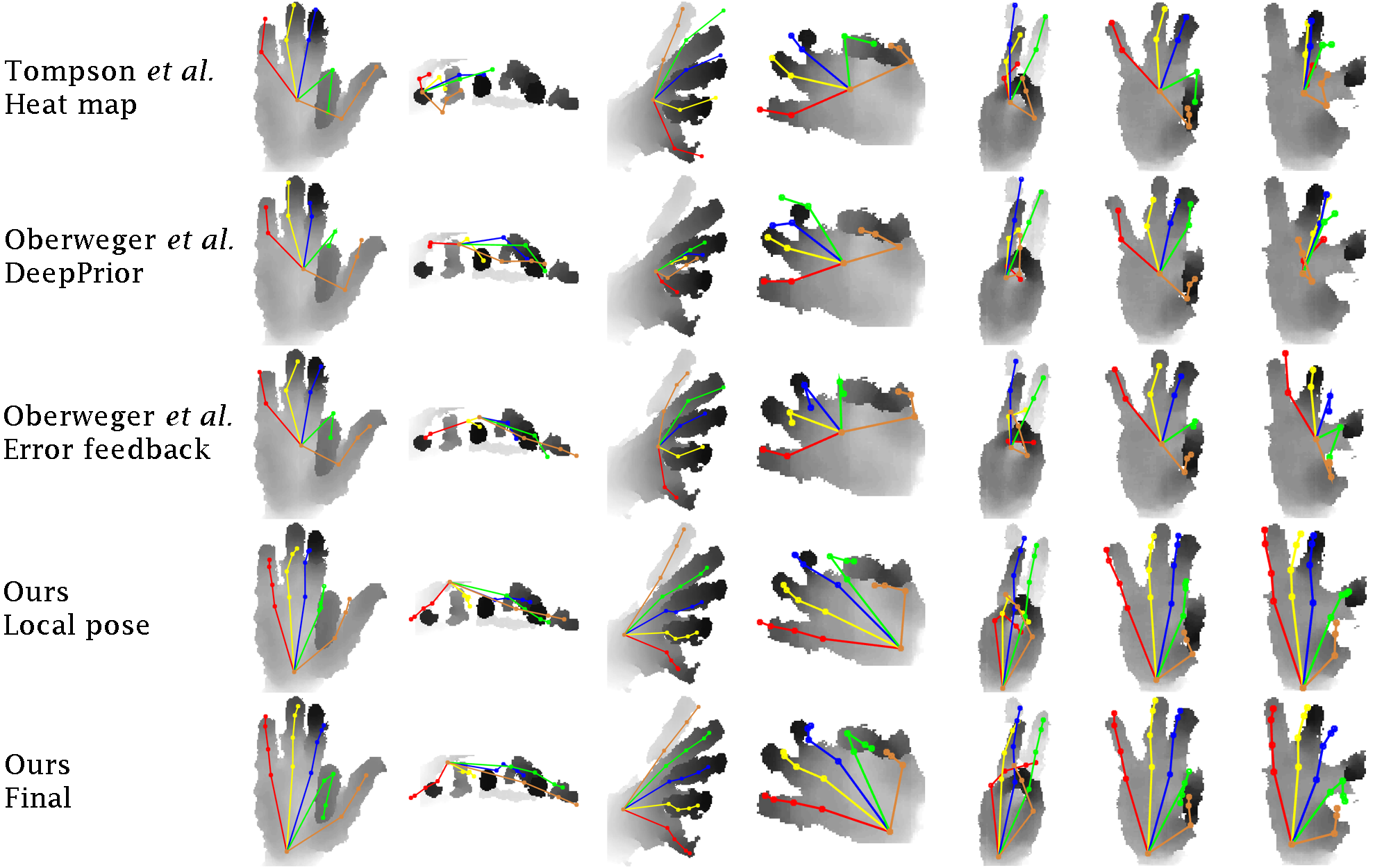}
  \caption[diagram]{Qualitative comparison of our proposed approach vs state-of-the-art methods.}
  \label{fig:diagram}\vspace{-0.4cm}
\end{figure}

In this paper, direct regression of the 3D hand pose is implemented as a specific tree-shaped CNN architecture designed to avoid training a coarse, global hand motion model, but allowing instead local finer specializations for different fingers and hand regions. So we break the hand pose estimation problem into hierarchical optimization subtasks, each one focused on a specific finger and hand region. Combined together in a tree-like structure, the final CNN shows fast convergence rates due to computations applied at a local level. In addition, we model correlated motion among fingers by fusing the features, learned in the hierarchy, through fully connected layers and training the whole network in an end-to-end fashion. The main advantage of this strategy is that the 3D hand pose prediction problem is attained as a global learning task based on local estimations. 

Moreover, it has been proved that $L2$ loss, in regression problems, is sensitive against outliers and ground-truth noise \cite{belagiannis2015robust}. Therefore, in order to further improve the final estimation in high non-linear spaces of hand configurations, we incorporate appearance and physical penalties in the loss function, based on the physical constraints typically applied in 3D reconstruction of human poses ~\cite{akhter2015pose}. By including such penalties during the network learning stage, unrealistic pose configurations are avoided. 

Lastly, as it is common in deep learning problems, variability and amount of data defines the success of a model and it has been proved that CNN models can not generalize well to unseen data. In this paper we introduce a non-rigid augmentation approach to generate realistic data from training data. To the best of our knowledge this is the first time such augmentation is applied in depth images. We use ground truth joints to compute hand kinematic parameters and deform hand joints. We then apply interpolation techniques to deform point cloud based joints. Results demonstrate that our proposed framework trained on augmented data outperforms state-of-the-art data-driven approaches in NYU and MSRA datasets.

We qualitatively compare pose estimation state-of-the-art approaches w.r.t. ours in Fig.~\ref{fig:diagram}. The work of Tompson \etal \cite{tompson2014} estimates 2D pose using joints heat-map only, thus providing poor pose estimation results in the case of noisy input images (second column). Oberweger \etal \cite{Oberweger2015} results (DeepPrior) show that PCA is not able to properly model hand pose configurations. Oberweger \etal \cite{Oberweger2015iccv} improved previous results by applying an error feedback loop approach. However, error feedbacks do not provide accurate pose recovery for all the variability of hand poses. In essence, in our proposed local-based pose estimation framework, a separate network is trained for each finger. Subsequently, we fuse such learned local features to include higher order dependencies among joints, thus obtaining better pose estimation results than previous approaches.

\section{Related Work}
\label{sec:related}

Hand pose estimation has been extensively studied in literature~\cite{Erol2007}, we refer the reader to \cite{Sharp2015} for a complete classification of state-of-the-art works in the field. Here we focus mostly on recent works using CNNs and depth cameras.

Most CNN-based architectures in data-driven hand pose estimation approaches are specifically designed to be discriminative and generalizable. Although the success of such approaches depends on the availability and variability of training data, CNN models cope reasonably well with this problem, and two main families of approaches can be distinguished in the literature, namely heat-map and direct regression methods.

Heat-map approaches estimate likelihoods of joints for each pixel as a pre-processing step. In \cite{tompson2014}, a CNN is fed with multi resolution input images and one heat-map per joint is generated. Subsequently, an inverse kinematic model is applied on such heat-maps to recover the hand pose. Nevertheless, this approach is prone to propagate errors when mapping to the original image, and estimated joints may not correlate with the hand physics constraints. The work of \cite{ge2016} extends this strategy by applying multi-view fusion of extracted heat-maps, where 3D joints are recovered from only three different viewpoints. In this approach, erroneous heat-maps are expected to be improved in the fusion step using complementary viewpoints. The key idea in this work is to reduce the complexity of input data by aligning all data with respect to the hand point cloud eigenvectors. For most heat-map based approaches, however, an end-to-end solution can be only achieved by considerably increasing the complexity of the model, e.g. introducing a cascading approach~\cite{wei2016}. Although such approaches used to work well for 2D pose estimation in RGB images, they are not necessarily able to model occluded joints from complex hand poses in depth data.

As an alternative, a number of works propose direct regression for estimating the joint positions of the 3D hand pose based on image features \cite{Oberweger2015iccv,Oberweger2015,Sun2015}. As mentioned in \cite{tompson2015efficient}, contrary to heat-map based methods, hand pose regression can better handle the increase in complexity of modeling highly nonlinear spaces. Although some approaches propose Principle Component Analysis (PCA) to reduce the pose space \cite{ge2016,Oberweger2015}, in general such linear methods typically fail when dealing with large pose and appearance variabilities produced by different viewpoints (as shown in Fig.~\ref{fig:diagram}).

Recently, error feedback~\cite{Oberweger2015iccv} and cascading~\cite{Sun2015} approaches have proven to avoid local minima by iterative error reduction. Authors in \cite{Oberweger2015iccv} propose to train a generative network of depth images by iteratively improving an initial guess. In this sense, Neverova \etal \cite{neverova2015hand} use hand segmentation as an intermediate representation to enrich pose estimation with iterative multi-task learning. Also, the method proposed in \cite{Sun2015} divides the global hand pose problem into local estimations of palm pose and finger poses. Thus, finger locations can be updated at each iteration relative to the hand palm. Contrary to our method, the authors use a cascade of classifiers to combine such local estimations. 

Authors in \cite{sinha2016deephand} apply a CNN to make use of the resulting feature maps as the descriptors for computing k-nearest shapes. Similarly to our approach, in their method the CNN separates palm and fingers and computes the final descriptor by dimensionality reduction. Differently to our approach, they factorize the feature vectors and nearest neighbors hyper-parameters to estimate the hand pose. In a different way, we propose training the network by fusing local features to avoid non-accurate local solutions, without the need of introducing cascading strategies nor multi-view set-ups. Contrary to the methods trying to simplify the problem by dividing the output space into subspaces, Guo \etal \cite{guo2017region} divided input image to smaller overlapping regions and fused CNN feature maps as a region ensemble network.

In CNN-based methods, data augmentation is a common approach to boost network to generalize better. Recently, Ge \etal \cite{ge20173d} applied data augmentation in the problem of hand pose recovery and showed a meaningful improvement in the results. Even Oberweger \etal \cite{oberweger2017deepprior++} extended DeepPrior model in \cite{Oberweger2015} and showed effectiveness of a simple model trained with data augmentation. However, aforementioned approaches use simple and rigid data augmentation like scaling, rotation and translation, which may not represent the visual variability in terms of 3D articulated joints. Here, we propose a non-rigid data augmentation by deforming hand parameters and interpolating point cloud.

\section{Global hand pose recovery from local estimations}
\label{sec:method}
Given an input depth image $\mathcal{I}$, we refer the 3D locations of $n$ hand joints as the set $J=\{ j\in\mathbb{R}^3\}_{1}^{n}$. We denote $j^{xyz}$ and $j^{uvz}$ as a given joint in the world coordinate system and after projecting it to the image plane, respectively. We define $n=20$ for the wrist, finger joints and finger tips, following the hand model defined in \cite{Sun2015}. We assume a hand is initially visible in the depth image, i.e. not occluded by other objects in the scene, although may present self-occlusions, and properly been detected beforehand (i.e. pixels belonging to the hand are already segmented~\cite{tompson2014}). We also assume intrinsic camera parameters are available. We refer to global pose as the whole set $J$, while, a local pose is a subset of $J$ (e.g. index finger joints).

Considering hand pose recovery as a regression problem with the estimated pose as output, 
we propose a CNN-based tree-shaped architecture, starting from the whole hand depth image and subsequently branching the CNN blocks until each local pose. We show the main components of the proposed approach in Fig. \ref{fig:arch}. In such a design, each network branch is specialized in each local pose, and related local poses share features in the earlier network layers. Indeed, we break global pose into a number of overlapping local poses and solve such simpler problems by reducing the nonlinearity of global pose. However, since local solutions can be easily trapped into local minima, we incorporate higher order dependencies among all joints by fusing the last convolutional layer features of each branch and train the network for global and local poses jointly. We cover this idea in Sec.~\ref{sec:cnn}.  
We also apply constraints based on appearance and dynamics of hand as a new effective loss function which is more robust against overfitting than simple $L2$ loss while providing a better generalization. This is explained in Sec.~\ref{sec:loss}.

\subsection{Hand pose estimation architecture}
\label{sec:cnn}
\begin{figure}[!t]
  \centering
  \includegraphics[width=\linewidth]{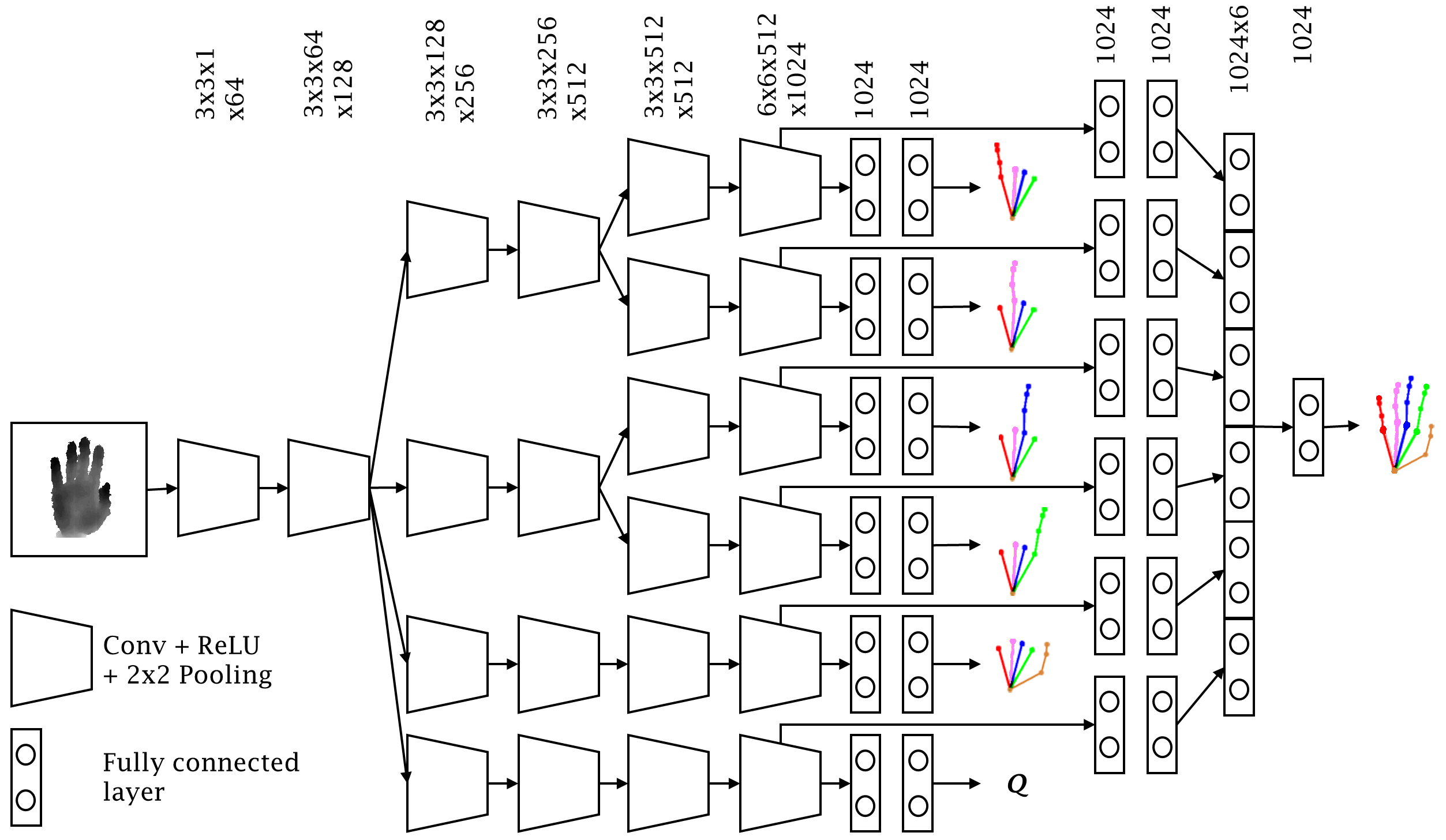}
  \caption[diagram]{Proposed network architecture. Branching strategy connects CNN blocks into a tree-shape structure while regressing local pose at each branch. Each local pose is a 24 dimensional vector. We also include a viewpoint regressor in th network as a rotation matrix in terms of quaternions $Q$ at the output. We then fuse all the features of the last convolutional layers to estimate output global pose. We use $Q$ features in the fusion to extract palm joints more accurate.}
  \label{fig:arch}\vspace{-0.4cm}
\end{figure}

In CNNs, generally, each filter extracts a feature from a previous layer, and by increasing the number of layers, a network is able to encode different inputs by growing the Field of View (FoV). During training, features are learned to be activated through a nonlinear function, for instance using Rectified Linear units (ReLU). The complexity and number of training data has a direct relation to the number of filters, layers or complexity of the architecture: an enormous number of filters or layers might cause overfitting, while a low number might lead to slow convergence and poor recognition rates. Interestingly, different architectures have been proposed to cope with these issues \cite{wei2016,Oberweger2015iccv,tompson2015efficient}. For example, in multi-task learning, different branching strategies are typically applied to solve subproblems~\cite{dosovitskiy2015learning,fan2015combining}, and the different subproblems are solved jointly by sharing features. Similarly, we divide global hand pose into simpler local poses (i.e. palm and fingers) and solve each local pose separately in a branch by means of a tree-shaped network. We show this architecture in Fig.~\ref{fig:arch}.

The proposed architecture has several advantages. Firstly, most correlated fingers share features in earlier layers. By doing this, we allow the network to hierarchically learn more specific features for each finger with respect to its most correlated fingers. Secondly, the number of filters per finger can be adaptively determined. Thirdly, the estimation of the global pose is reduced to the estimations of simpler local poses, resulting the network to train at fast convergence rates.

We define the amount of locality by the number of joints contributing to a local pose. Keeping such locality high (i.e. lower number of joints), in one hand, causes fingers to be easily confused among each other, or detected in a physically impossible location. A low locality value (i.e. higher number of joints), on the other hand, increases the complexity. Besides, local joints should share a similar motion pattern to keep lower complexity. So in the particular implementation in this paper, we assign to each local pose one finger plus palm joints, thus leading to a 24 dimensional vector.

Training the network only based on local poses omits information about inter-fingers relations. Tompson \etal~\cite{Tompson2014nips} included a graphical model within the training process to formulate joints relationships. Li \etal~\cite{Li2015} used a dot product to compute similarities for embedded spaces of a given pose and an estimated one in a structural learning strategy. Instead, we apply late fusion based on local features, thus, let the network learn the joint dependencies through fully connected layers for estimating the final global pose. The whole network is trained end-to-end jointly for all global and local poses given a constrained loss function.

\textbf{Network details.} Input images are pre-processed with a fix-sized cube centered on the hand point cloud and projected into the image plane. Subsequently, the resulting window is cropped and resized to a $192\times192$ fixed size image using nearest neighbor interpolation, with zero-mean depth.

As intermediate layers, the network is composed of six {\em branches}, where each branch is associated to specific fingers as follows: two branches for index and middle fingers, two branches for ring and pinky fingers, one branch for thumb, and one branch for palm. For the palm branch, instead of performing direct regression on palm joints, we make regression on the palm viewpoint, defined as the rotation (in terms of quaternions) between the global reference view and the palm view. As shown in the experimental results, more accurate and reliable optimization is then achieved, since the network is able to model interpolations among different views.

As shown in Fig.~\ref{fig:arch}, each convolutional block consists of a convolution layer with $3\times3$ filter kernels and a ReLU followed by a max-pooling, except for the last block. All pooling layers contain a $2\times2$ window. The last block contains a convolutional layer with $6\times6$ filter kernels, providing a feature vector. Fully connected layers are added to the end of each branch for both local and global pose learning. For local pose at each branch there are two hidden layers with 1024 neurons with a dropout layer in between. Similarly, for global pose at each branch, the feature vector is followed by two hidden layers with 1024 neurons with a dropout layer in between. Then, the last hidden layers are concatenated and followed by a dropout and a hidden layer with 1024 neurons. Finally, the global and local output layers provide the estimation of joints with one neuron per joint and dimension.

\subsection{Constraints as loss function}
\label{sec:loss}
In regression problems, the goal is to optimize parameters such that a loss function between the estimated values of the network and the ground-truth value is getting minimized. Usually, in the training procedure, an $L2$ loss function plus a regularization term is optimized. However, it is generally known that, in an unbalanced dataset with availability of outliers, $L2$ norm minimization can result in poor generalization and sensitivity to outliers where equal weights are given to the training data~\cite{belagiannis2015robust}. Weight regularization is commonly used in deep learning as a way to avoid overfitting. However, it does not guarantee the weight updating to bypass the local minima. Besides, a high weight decay causes low convergence rates. Belagiannis \etal~\cite{belagiannis2015robust} proposed Tukey’s biweight loss function in the regression problems as an alternative to $L2$ loss robust against outliers. We formulate the loss function as $L2$ loss along with constraints applied to hand joints regarding the hand dynamics and appearance, leading to more accurate results and less sensitivity to ground-truth noise.
 We define the loss function for one frame in the form of:
\begin{equation}
L=\lambda_1 L_{loc} + \lambda_2 L_{glo} + \lambda_3 L_{app} + \lambda_4 L_{dyn},
\label{eq:loss}
\end{equation}
where $\lambda_i$ $i\in\{1..4\}$ are factors to balance loss functions. $L_{loc}$, $L_{glo}$, $L_{app}$ and $L_{dyn}$ denote the loss for the estimated local and global pose, appearance, and hand dynamics, respectively. Next, each component is explained in detail.

Let $F^l\in\mathbb{R}^{3 \times m}$ be the concatenation of the $m$ estimated joints in each branch of the proposed network and $G^l\in\mathbb{R}^{3 \times m}$ be the ground-truth matrix. Note that $m$ is not necessarily equal to $n=20$. $F^{g}\in\mathbb{R}^{3 \times n}$ and $G^{g}\in\mathbb{R}^{3 \times n}$ are the outputs of the embedded network for estimated joints and ground-truth, respectively. Then, we define local and global losses as:
\begin{equation}
L_{loc}=\sum_{i=1}^{3m} (F^l_i-G^l_i)^2,
\label{eq:loss_j}
\end{equation}
\begin{equation}
L_{glo}=\sum_{i=1}^{3n} (F_{i}^{g}-G_{i}^{g})^2.
\label{eq:loss_e}
\end{equation}

A common problem in CNN-based methods for pose estimation is that in some situations estimated pose does not properly fit with appearance. For instance, joints are placed in locations where there is no evidence of presence of hand points, or being physically incorrect \cite{Oberweger2015iccv,Oberweger2015, ge2016}. In this paper, during training we penalize those joint estimations that do not fit with the appearance or are physically not possible, and include such penalties in the loss function. 

We first assume that, rationally, joints must locate inside the hand area and have a depth value higher than the hand surface, besides, joints must present physically possible angles in the kinematic tree. Therefore, for a given joint $j^{xyz}$ the inequality $\mathcal{I}(j^u,j^v)-j^z<0$ must hold, where $\mathcal{I}(j^u,j^v)$ is the pixel value at location $(j^u,j^v)$. To avoid violating the first condition (i.e. when a joint is located outside hand area after projection to the image plane), we set the background with a cone function as: $$5\sqrt{(u-0.5w)^2+(v-0.5h)^2}+\phi,$$ where $w$ and $h$ are width and height of the image, and $\phi$ is a fixed value set to 100. The reason to use a cone function instead of a fixed large value is to avoid zero derivatives on the background. We use hinge formulation to convert inequality to a loss through:
\begin{equation}
L_{app}=\sum_{i=1}^{m}{\max(0,\mathcal{I}(j_{i}^{u},j_{i}^{v})-j_{i}^{z})}.
\label{eq:loss_a}
\end{equation}

  
  
  

We subsequently incorporate hand dynamics by means of the top-down strategy described in Algorithm~1. We assume all joints belonging to each finger (except thumb) should be collinear or coplanar. Thumb has an extra non-coplanar form and we do not consider it in the hand dynamics loss. A groundtruth finger state $s_G\in \{1..4\}$ is assigned to each finger computed by the conditions defined in Algorithm~1. Each finger has a groundtruth normal vector $\mathbf{e}_G$ which is finger direction for the case~1 and finger plane normal vector for the other cases. Therefore, we define four different losses, one of them triggered for each finger (as shown in Algorithm~1). Let $A$, $B$, $C$ and $D$ be four joints belonging to a finger starting in $A$ as the root joint and ending in $D$ as fingertip. Then the dynamics loss is defined as:
\begin{equation}
L_{dyn}=\sum_{i=1}^{4}\Delta_i(A,B,C,D,s_G,\mathbf{e}_G),
\end{equation}
where $i$ denotes a finger index. Now we consider each case in Algorithm~1 in the following.

We consider a collinear finger in case~1. A finger is collinear if:$$\|B-A\|+\|C-B\|+\|D-C\|<\|D-A\|+\kappa,$$ 
where $\kappa$ is a threshold defining the amount of collinearity and set to $0.01\|D-A\|$. To compute the loss for a collinear groundtruth finger, the following condition has to be hold: $\rho<\cos{(\angle (\overrightarrow{AD}, \mathbf{e}_G))}\leq1,$ 
where $\rho$ is a threshold. This condition has to be met for $\overrightarrow{AB}$ and $\overrightarrow{AC}$ as well. The cosine function can be extracted through dot product. Therefore, using hinge formulation, the loss is defined as:
\begin{equation}
\begin{aligned}
 \Delta_i & (A,B,C,D,1,\mathbf{e}_G)= \\ \label{eq:loss_l}
 & \max\left(0,\rho-\frac{\overrightarrow{AB}\cdot\mathbf{e}_G}{\|\overrightarrow{AB}\|}\right) +\\
 & \max\left(0,\rho-\frac{\overrightarrow{AC}\cdot\mathbf{e}_G}{\|\overrightarrow{AC}\|}\right) +\\
 & \max\left(0,\rho-\frac{\overrightarrow{AD}\cdot\mathbf{e}_G}{\|\overrightarrow{AD}\|}\right) +\\
 & \mu\max(0,\|\overrightarrow{AB}\|+\|\overrightarrow{BC}\|+\|\overrightarrow{CD}\|-1.01\|\overrightarrow{AD}\|),
\end{aligned}
\end{equation}
where $\mu$ is a factor to balance different components of the loss function.

\begin{figure}[!t]
  \centering
  \includegraphics[width=\linewidth]{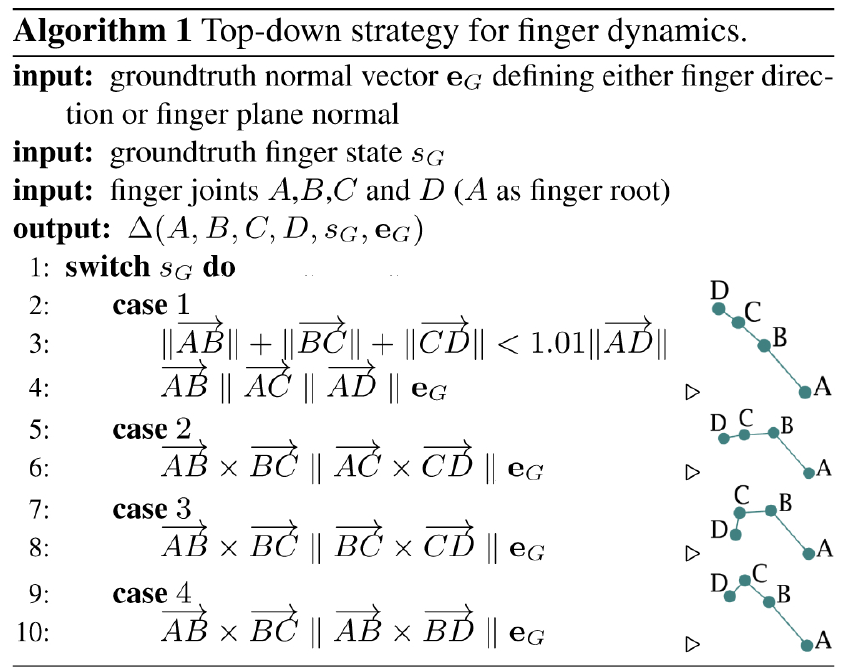}
  \label{fig:dynamics}\vspace{-.6cm}
\end{figure}
We consider a coplanar finger for cases~1,~2~and~3. We define a finger to be coplanar if cross products of all subsets of the finger joints with three members to be parallel. Note that a collinear finger is necessarily coplanar. However, we exclude collinear fingers from this definition due to cross product, as shown in Algorithm~1. For a groundtruth coplanar finger, such cross products must be parallel to the plane normal vector. Therefore, for given joints $A$, $B$ and $C$, the following condition must hold: $$\rho<\cos{(\angle (\overrightarrow{AB} \times \overrightarrow{BC},\mathbf{e}_G)}\leq1.$$ Given the groundtruth finger is coplanar of case $2$, we compute the loss function as:
\begin{equation}
\begin{aligned}
\Delta_i & \resizebox{.9\hsize}{!}{ $\displaystyle{(A,B,C,D,2,\mathbf{e}_G)=\max\left(0,\rho-\frac{(\overrightarrow{AB}\times\overrightarrow{BC})\cdot\mathbf{e}_G}{\|\overrightarrow{AB}\times\overrightarrow{BC}\|}\right)}$}  \\
 & +\max\left(0,\rho-\frac{(\overrightarrow{AC}\times\overrightarrow{CD})\cdot\mathbf{e}_G}{\|\overrightarrow{AC}\times\overrightarrow{CD}\|}\right).
\end{aligned}
\label{eq:loss_p}
\end{equation}
The loss functions for the other coplanar finger cases are computed in the same way.

\subsection{Loss function derivatives}
\label{sec:derivatives}
All components in Eq. \ref{eq:loss} are differentiable, thus we are able to use gradient-based optimization methods. In this section we explain derivatives of the constraint loss function in Eq. \ref{eq:loss_a}. Derivatives of the rest of loss functions are computed through matrix calculations. We first define derivative of $L_{app}$ with respect to $t\in\{j_{i}^{x},j_{i}^{y},j_{i}^{z}\}$ through:

\begin{equation}
 \frac{\partial L_{app}}{\partial t}=
 \begin{cases}
    0       & \quad \text{if } \mathcal{I}(j_{i}^{u},j_{i}^{v})-j_{i}^{z}\leq0\\
    {\partial \mathcal{I}}/{\partial t} - {\partial j_{i}^{z}}/{\partial t} & \quad \text{otherwise}. \\
  \end{cases}
\label{eq:der_loss_a}
\end{equation}
In the following we just consider positive condition of Eq. \ref{eq:der_loss_a}. Besides, we omit index $i$ (which denotes $i$-th joint) from the notations for the easiness of reading. Depth image $\mathcal{I}$ is a discrete multi-variable function of $j^{u}$ and $j^{v}$, where $j^{u}$ is a multi-variable function of $j^{x}$ and $j^{z}$, and $j^{v}$ is a multi-variable function of $j^{y}$ and $j^{z}$. Consequently, the total derivative of a depth image can be computed by the chain rule through:
\begin{equation}
\frac{\mathrm{d} \mathcal{I}}{\mathrm{d} t} = \frac{\partial \mathcal{I}}{\partial j^{u}} \frac{\mathrm{d} j^{u}}{\mathrm{d} t} + \frac{\partial \mathcal{I}}{\partial j^{v}} \frac{\mathrm{d} j^{v}}{\mathrm{d} t}
\end{equation}
\begin{equation}
\frac{\mathrm{d} j^{u}}{\mathrm{d} t} = \frac{\partial j^{u}}{\partial j^{x}} \frac{\mathrm{d} j^{x}}{\mathrm{d} t} + \frac{\partial j^{u}}{\partial j^{z}} \frac{\mathrm{d} j^{z}}{\mathrm{d} t}
\label{eq:dudt}
\end{equation}
\begin{equation}
\frac{\mathrm{d} j^{v}}{\mathrm{d} t} = \frac{\partial j^{v}}{\partial j^{y}} \frac{\mathrm{d} j^{y}}{\mathrm{d} t} + \frac{\partial j^{v}}{\partial j^{z}} \frac{\mathrm{d} j^{z}}{\mathrm{d} t}
\end{equation}

Next, we present components of $j^{u}$ derivative in detail\footnote{Derivatives belonging to $j^{v}$ are computed in the same way as $j^{u}$}. Depth image $\mathcal{I}$ is a function of hand surface. However, hand surface given by the depth camera may have noise and not be differentiable at some points. To cope with this problem, we estimate depth image derivatives by applying hand surface normal vectors. Let $\mathbf{s}$ to be the surface normal vector for a given joint. Then, derivative of $\mathcal{I}$ with respect to $u$ axis is given by the tangent vectors through:
\begin{equation}
\frac{\partial \mathcal{I}}{\partial j^{u}} = \frac{\mathbf{s}^{x}}{\mathbf{s}^{z}}.
\end{equation}

As mentioned, $j^{uvz}$ is the projection of the estimated joint $j^{xyz}$ from world coordinate to the image plane. Note that joints have zero mean and $j^{uvz}$ is extracted after the image has been cropped and resized. Let $f_x$, $p_x$, $M^{xyz}$ and $M^{uvz}$ to be the camera focal length and image center for $x$ axis, world coordinate hand point cloud center, and its projection to image plane, respectively. Then $j^{u}$ is computed as:
\begin{equation}
\begin{aligned}
& j^{u}(j^x,j^z)=\left( \frac{f_x(j^x+M^x)}{j^z+M^z}+p_x-M^u\right)\text{scale}_x+\frac{w}{2},\\
& \text{scale}_x=\frac{wM^z}{cf_x},
\end{aligned}
\end{equation}
where $c$ is the cube size used around hand point cloud to crop the hand image. Using this formulation, derivative of $j^{u}$ can be easily computed and replaced in Eq.~\ref{eq:dudt}.

\section{Experiments}
\label{sec:experiments}
In this section we evaluate our approach on two real-world datasets NYU \cite{tompson2014} and MSRA \cite{Sun2015}, and one synthetic dataset SyntheticHand \cite{madadiFG2017}. NYU dataset has around 73K annotated frames as training data (single subject) and 8K frames as test data (two subjects). Each frame has been captured from 3 different viewpoints and ground truth is almost accurate. MSRA dataset has 76K frames captured from 9 subjects each in 17 pose categories. This dataset does not provide an explicit training/test set and a subject exclusive 9-fold cross validation is used to train and evaluate on this dataset. MSRA dataset has smaller image resolution and less pose diversity and accurate ground truth comparing to NYU dataset. SyntheticHand dataset has over 700K training data and 8K test data consisting of a single synthetic subject performing random poses from all viewpoints, thus being useful to analyze our methodology under occlusions. All three datasets provide at least 20 hand joints in common. However, NYU dataset has 16 extra joints. We evaluate our approach using two metrics: average distance error in mm and success rate error~\cite{taylor2012vitruvian}. Next, we detail the method parameters and evaluate our approach both quantitatively and qualitatively in comparison to state-of-the-art alternatives.

\begin{figure}[!t]
\centering
\includegraphics[width=0.9\linewidth]{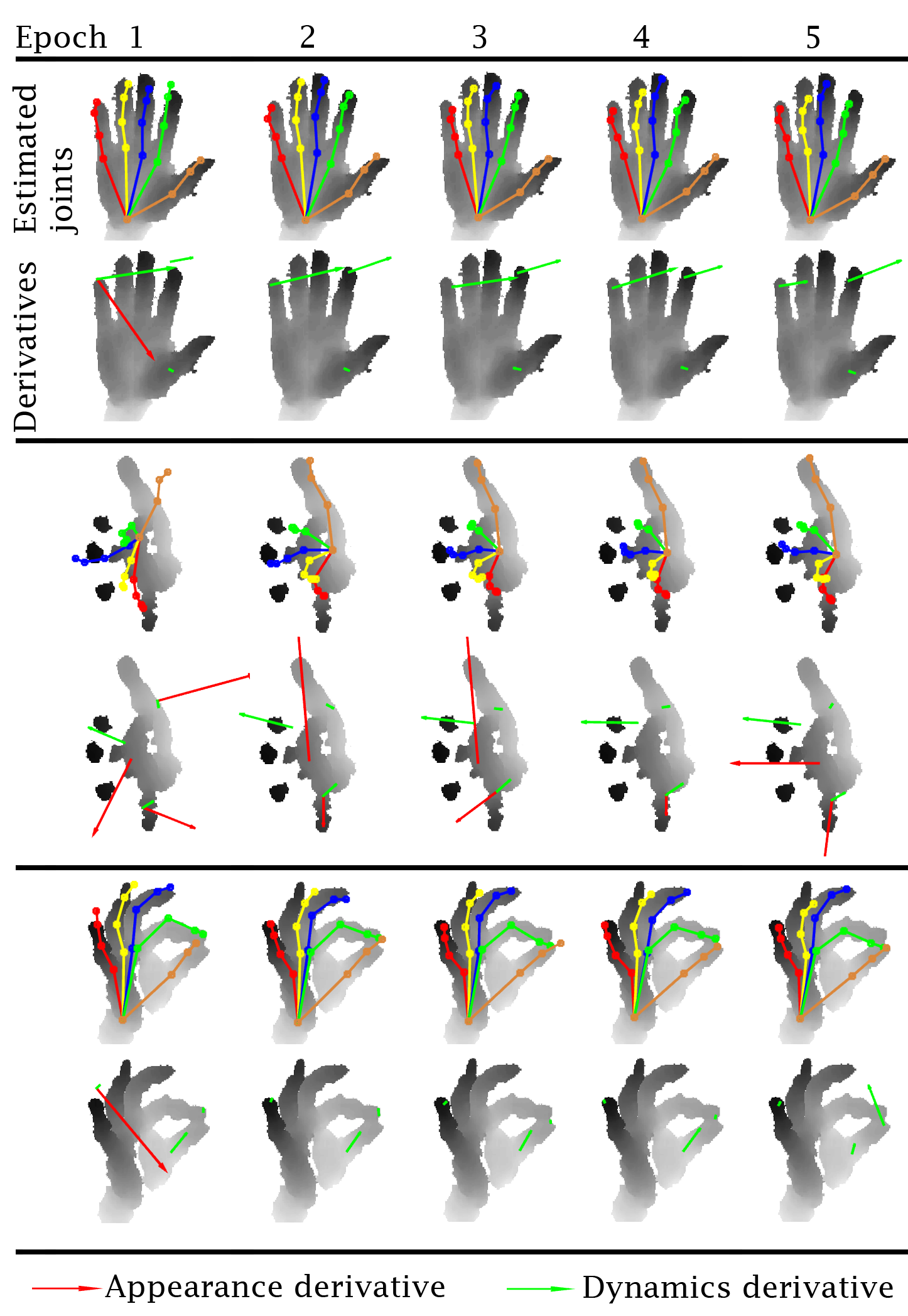}
  \label{fig:derivative}
  \caption{Constraints derivatives during training process (original NYU dataset). Estimated joints along with derivatives of appearance and hand dynamics are illustrated for the first five epochs in the training process. We qualitatively show how proposed network converges very fast in few epochs.}\vspace{-0.6cm}
\end{figure}

\subsection{Training}
\label{sec:training}
We utilize MatConvNet library \cite{vedaldi15matconvnet} on a server with GPU device \textit{GeForce GTX Titan X} with 12 GB memory. We optimize the network using stochastic gradient descent (SGD) algorithm. We report hyper-parameters used in NYU dataset. We set the batch size, learning rate, weight decay and momentum to 50, 0.5e-6, 0.0005 and 0.9, respectively. Our approach converges in almost 6 epochs while reducing the learning rate by a factor of 10 for two more epochs. Overall, training takes two days on original NYU dataset while testing takes 50 fps.


\begin{figure}[!t]
\centering
\begin{subfigure}[]
  {\includegraphics[width=\linewidth]{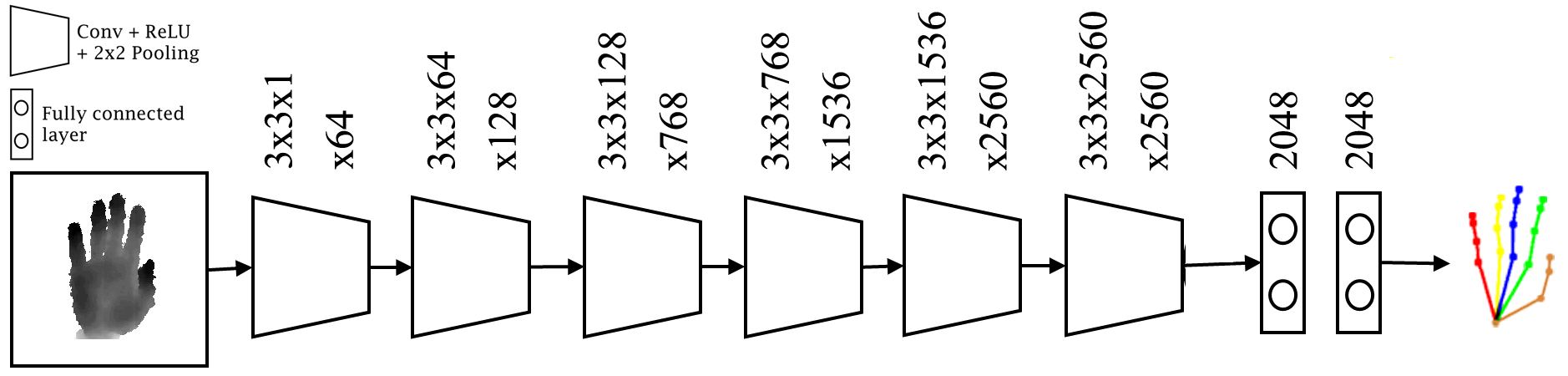}
  \label{fig:network-single-channel}}
  \end{subfigure}
  \begin{subfigure}[]
  {\includegraphics[width=\linewidth]{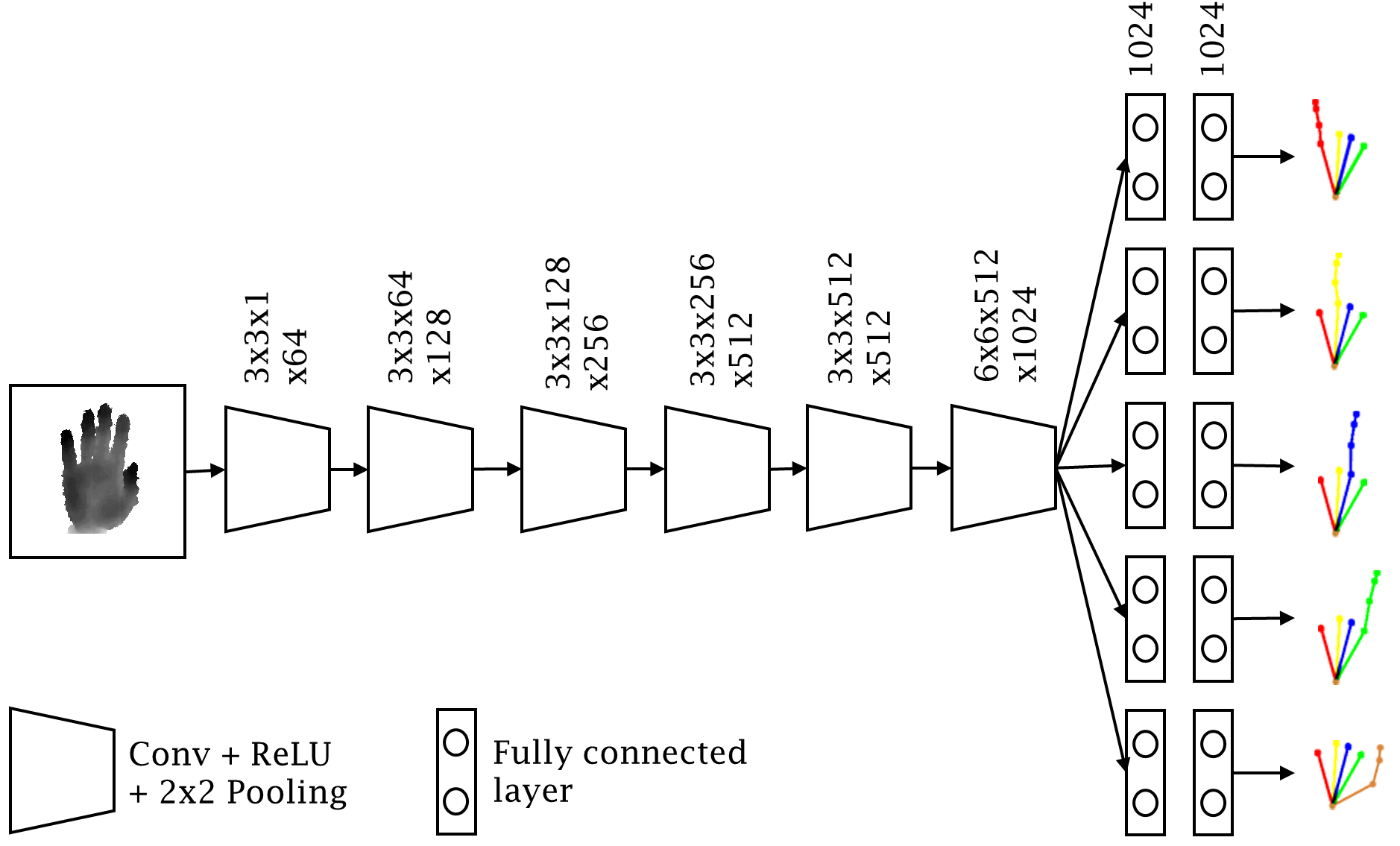}
  \label{fig:network-fc-branching}}
  \end{subfigure}
\caption{Baseline architectures. a) A single channel network with the same convolutional capacity as branching network. We train this network with the same loss as in Eq. \ref{eq:loss} omitting $L_{loc}$. b) A single channel network with the same capacity as one branch in hierarchical model and branching local pose on FC layers. We train this network with the same loss as in Eq. \ref{eq:loss} omitting $L_{glo}$.}
  \label{fig:network-baseline}
\end{figure} 

\begin{figure*}[!t]
  \centering
  \begin{subfigure}[]
  {\includegraphics[width=0.45\linewidth,height=4.8cm]{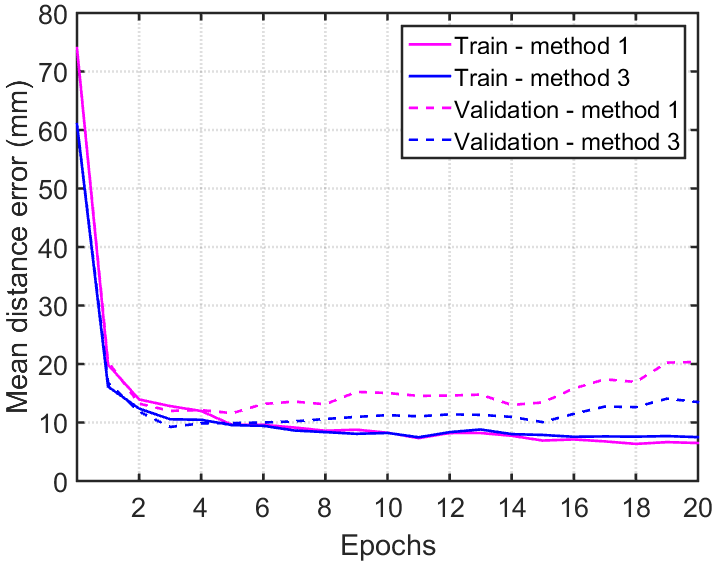}
  \label{fig:train-trend}}  
  \end{subfigure}
  \begin{subfigure}[]
  {\includegraphics[width=0.45\linewidth,height=4.8cm]{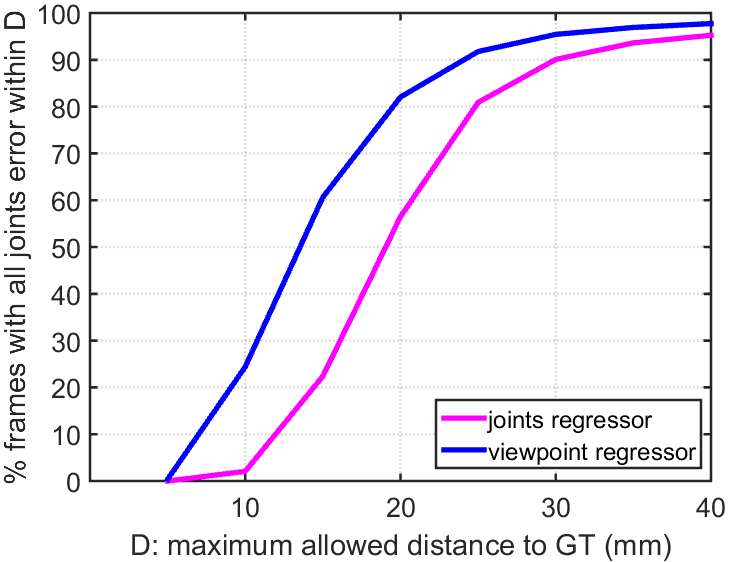}
  \label{fig:palm-joints-vs-view}}  
  \end{subfigure}\\  
  \begin{subfigure}[]
  {\includegraphics[width=0.45\linewidth,height=4.8cm]{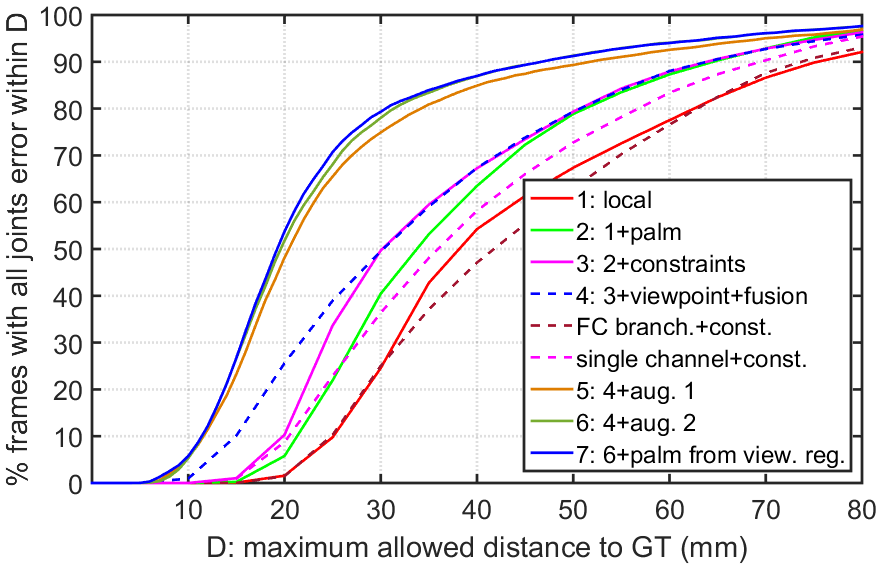}
  \label{fig:baseline}}  
  \end{subfigure}
  \begin{subfigure}[]
  {\includegraphics[width=0.45\linewidth,height=4.8cm]{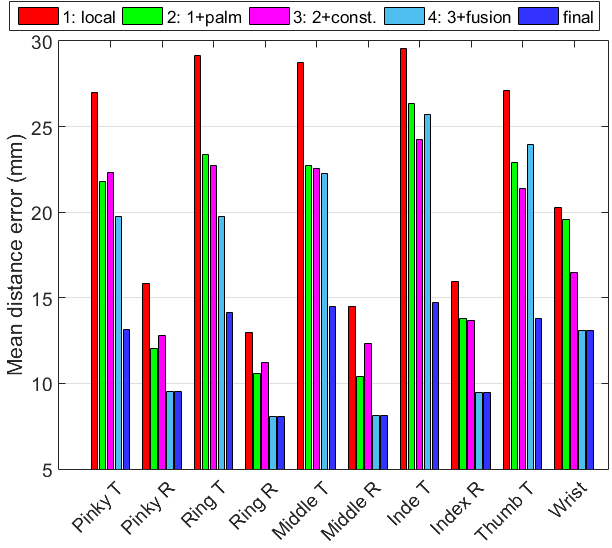}
  \label{fig:baseline-mean}}  
  \end{subfigure}
  \caption{Quantitative results comparing baselines on NYU dataset. a) Training process in terms of average error per epoch. b) Comparing palm: joints regression vs. viewpoint regression. c) Maximum error success rate comparing baselines. d) Per joint average error comparing baselines.}
\end{figure*}

\textbf{Loss function parameters tuning} 
We set a low value for parameter $\mu$ in Eq. \ref{eq:loss_l} since it behaves like a regularization and it is not connected to ground-truth. $L_{dyn}$ is mainly a summation of cosine functions while $L_{app}$ is in millimeters. Therefore we set $\lambda_4$ higher than $\lambda_3$ to balance cosine space with millimeter. Finally, we set parameters $\lambda_1$, $\lambda_2$, $\lambda_3$, $\lambda_4$ and $\mu$ experimentally to 4, 4, 3, 20 and 0.0005, respectively. We show derivatives of appearance and dynamics loss functions for a number of joints in the first five epochs in Fig.~3
, as well as qualitative images of estimated joints.

\begin{figure*}
\centering
\includegraphics[width=0.95\linewidth]{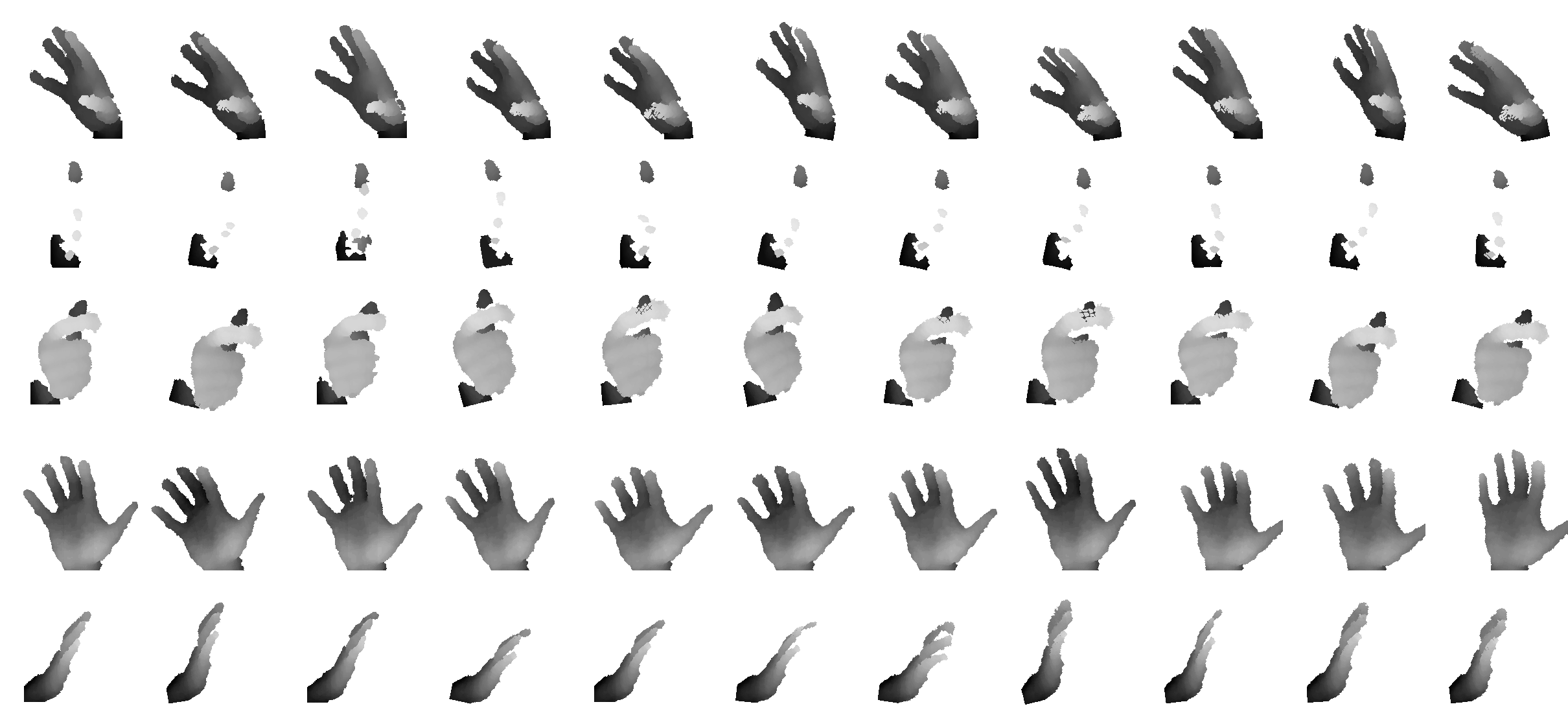}
\caption{Data augmentation. We generate new data by applying non-rigid hand shape deformation along with rigid transformations like in-plane rotation. Hand kinematic parameters are slightly deformed and new hand joints are used to interpolate hand point cloud. We also change palm and fingers size. Therefore, given a pose, different hand shapes can be generated which helps to generalize better to unseen subjects. First column shows the original images and others are generated samples.}
  \label{fig:data-aug}
\end{figure*}

\subsection{Ablation study}
In this section we study different components of the proposed architecture trained on NYU dataset. We denote each component by a number.

\textbf{Locality.} Locality is referred to the number of joints in the network output. In the first case, we analyze the hierarchical network trained just with one finger in each branch and without constraints and fusion network (so called \textit{1:local}). This network shows a high locality value. As one can expect, this network can easily overfit on the training data and exchange estimations for similar fingers. We show a significant improvement by decreasing locality by including palm joints in each branch (so called \textit{2:1+palm}). Palm joints are located in a near planar space and thus do not add high non-linearity to the output of each branch while help for better finger localization. We compare these methods in Fig. \ref{fig:baseline} (red vs. green lines).

\textbf{Constraints.} We train method \textit{2:1+palm} by including constraints in the loss function $L$ (Eq. \ref{eq:loss}) without $L_{glo}$ in this stage (so called \textit{3:2+constraint}). We still do not explicitly model any relationship among fingers in the output space, but let the network learn each finger joints with respect to the hand surface and finger dynamics. In Fig. \ref{fig:baseline} we show the effectiveness of this strategy (magnet line) against method \textit{2:1+palm}. We also analyze the effect of constraints in the training process in Fig. \ref{fig:train-trend}. As it can be seen, by applying the proposed constraints, method \textit{3:2+constraint} is more robust against overfitting than method 1. Validation error in method 3 does not significantly change from epoch 7 to 15. Comparing both methods in epoch 20, method 1 has a lower error in training while its validation error is almost 1.5 times the validation error of method 3.

\textbf{Branching strategy vs. single channel architecture.} As baselines, we created two single channel networks with 6 convolutional layers, as shown in Fig. \ref{fig:network-baseline}. The output of the first network (so called \textit{single-channel} network) is 3D locations of the full set of joints. In this architecture, the capacity of convolutional layers are kept similar to the whole branching network. This network is trained with loss function $L$ without $L_{loc}$. The outputs of the second network (so called \textit{FC-branching} network) is similar to the method \textit{3:2+constraint}. The capacity  of convolutional layers in this architecture is similar to one branch in tree-structure network. The branching in this network is applied on FC layers. We train this network with the same loss as method \textit{3:2+constraint}. We train both networks with the same hyper-parameters introduced in Sec. \ref{sec:training}. As one can see in Fig. \ref{fig:baseline}, single-channel network (dashed magnet line) performs worse than method \textit{3:2+constraint}, showing the effectiveness of the tree-structure network. This means regardless of the capacity of the network, in a single channel network, backpropagation of the gradients of the loss is not able to train network filters to map input image to a highly non-linear space in an optimal and generalizable solution. This is even worse for FC-branching network (dashed dark brown line).

\textbf{Palm viewpoint vs. palm joints regression.} We evaluate our palm joints vs. palm viewpoint regression in terms of success rate error in Fig. \ref{fig:palm-joints-vs-view}. Palm viewpoint regressor gives a rotation matrix in terms of quaternions. We convert quaternions to rotation matrix and use it to transform a predefined reference palm example. As it can be seen in the figure, palm viewpoint regression significantly reduces palm joints error.

\textbf{Global vs. local pose.} We add fusion network to method \textit{3:2+constraint} to model correlations among different local poses in an explicit way (so called \textit{4:3+viewpoint+fusion}). We include viewpoint regression features in the fusion as well. We illustrate the results in Fig. \ref{fig:baseline} (dashed blue line). Comparing to method \textit{3:2+constraint}, method \textit{4:3+viewpoint+fusion} improves performance for error thresholds below 30mm. 

\textbf{Per joint mean error.} We also illustrate per joint mean error in Fig.~\ref{fig:baseline-mean}. From the figure, as expected, a very local solution (method 1) performed the worst among the baselines. Comparing method 2 and \textit{3:2+constraint} in average error shows the benefits of applying constraints as loss, as well. By including viewpoint features in the fusion network, palm joints mean error was considerably reduced by method \textit{4:3+viewpoint+fusion}. Although method \textit{4:3+viewpoint+fusion} performed better for the pinky and ring fingertips, it did not achieve the best results for index and thumb fingertips.

\textbf{Data augmentation.} data augmentation is a common approach to boost CNN models with small deformations in the images. Mainly used data augmentation approaches are rotation, scaling, stretching and adding random noise to pixels. Such approaches are mainly rigid (rotation and scaling) or unrealistic (stretching). Here, we propose a realistic non-rigid data augmentation. As the first step, we remove redundant data by checking ground truth joints. In this sense a redundant data is an image which has a high similarity to at least one image in the training set. Such similarity is defined by maximum Euclidean distance $\Psi$ among corresponding joints. Therefore, two images are similar if $\Psi$ is below a threshold. We used threshold 10~mm for this task.

Our data augmentation is consisting of in-plane rotation, changing palm and fingers size and deforming fingers pose. We show some generated images in Fig. \ref{fig:data-aug}. In the following we explain details of data augmentation.

\begin{figure}[!ht]
\centering
\includegraphics[width=0.7\linewidth]{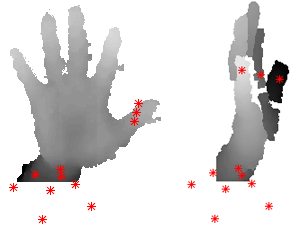}
  \caption{Auxiliary points '*' are added to the set of joints to avoid unrealistic warping in non-rigid hand augmentation.}
  \label{fig:aux-joints}
\end{figure}

The main idea in non-rigid hand deformation is to deform ground truth hand joints and interpolate point cloud based on new joints. We use thin plate spline (TPS) \cite{bookstein1989principal} as a standard interpolation technique to deform point cloud. However, to avoid extrapolation problem and unrealistic warping, we add some auxiliary points to the set of joints. We show some possible auxiliary points in Fig. \ref{fig:aux-joints}: we mainly add points around wrist and thumb. We observed unrealistic deformation around thumb and by adding three fixed points we avoided extrapolation problems. For the wrist case we do not want to deform points of lower arm. Fixed auxiliary points around wrist add constraint to space avoiding unrealistic warping.

A first possible shape deformation is the changing of hand scale. However, a simple scaling does not guarantee generalization to unseen subjects. Instead, we change the size of fingers and palm. This can be seen in the Fig. \ref{fig:data-aug} 4th row. As the first step we compute hand kinematic parameters based on hand coordinate. Hand coordinate is defined by palm joints such that, in a quite open hand, thumb defines $x$ coordinate direction, other fingers define $y$ coordinate direction and $z$ coordinate is perpendicular to palm plane. Then, palm can be stretched in the direction of $x$ or $y$. We stretch each direction by a random factor. Having kinematic parameters fixed, we are able to randomly modify fingers length and reconstruct new joints for each finger. It is also likely to slightly modify kinematic parameters and reconstruct joints in a new pose. However, we keep kinematic parameters near to original values to avoid unrealistic point cloud deformations and possible big holes in the depth image. Finally we apply morphological operations to fill small gaps. \footnote{Code will be publicly available after publication.}

In NYU dataset, around 60K images were remained after removing redundant images from all 218K samples in the training set (including all cameras). We then generated two sets of augmented images including around 780K and 1500K. We use random scaling factors in the range $[0.85,1.05]$ for palm and fingers. Kinematic parameters are changed by summation to random degrees in the range $[-7.5,7.5]$. The only difference in generated sets is the in-plane rotation degrees. The first and second sets have in-plane rotation in the range $[-30,30]$ and $[-90,90]$ degrees, respectively. 

We compare the results on both generated sets in Fig. \ref{fig:baseline} (brown and dark green lines). We train method \textit{4:3+viewpoint+fusion} on these new 2 sets, so called \textit{method 5:4+aug1} and \textit{method 6:4+aug2}, respectively. One can see the model trained on the set with more samples and wide in-plane rotation degrees (\textit{method 6:4+aug2}) generalizes better to the test set. Also, a significant improvement is achieved comparing to the original data (method \textit{4:3+viewpoint+fusion}). We observed that wrist joint has the maximum error in 20\% of the cases in method \textit{6:4+aug2}. Therefore we replaced estimated palm joints in method \textit{6:4+aug2} with estimated palm joints from viewpoint regressor and slightly improved performance (final method \textit{7:6+palm}). We also illustrate method \textit{7:6+palm} per joint mean error in Fig. \ref{fig:baseline-mean}. Fingertips have the highest error among the joints. Data augmentation helps to significantly improve the fingertip estimation, as we can see in Fig. \ref{fig:qualitative-nyu} comparing different baselines qualitatively.

\begin{figure*}[!ht]
  \centering
  \begin{subfigure}[]
  {\includegraphics[width=0.45\linewidth,height=4.8cm]{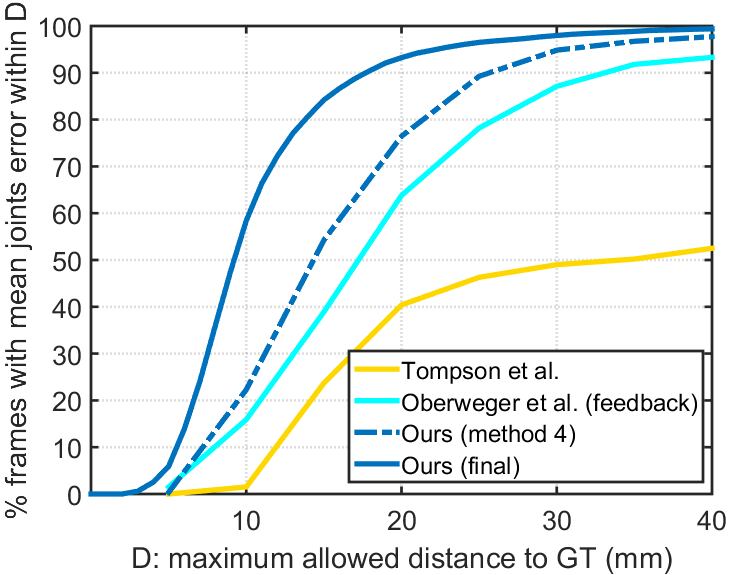}
  \label{fig:nyu-mean}}
  \end{subfigure}
  \begin{subfigure}[]
  {\includegraphics[width=0.45\linewidth,height=4.8cm]{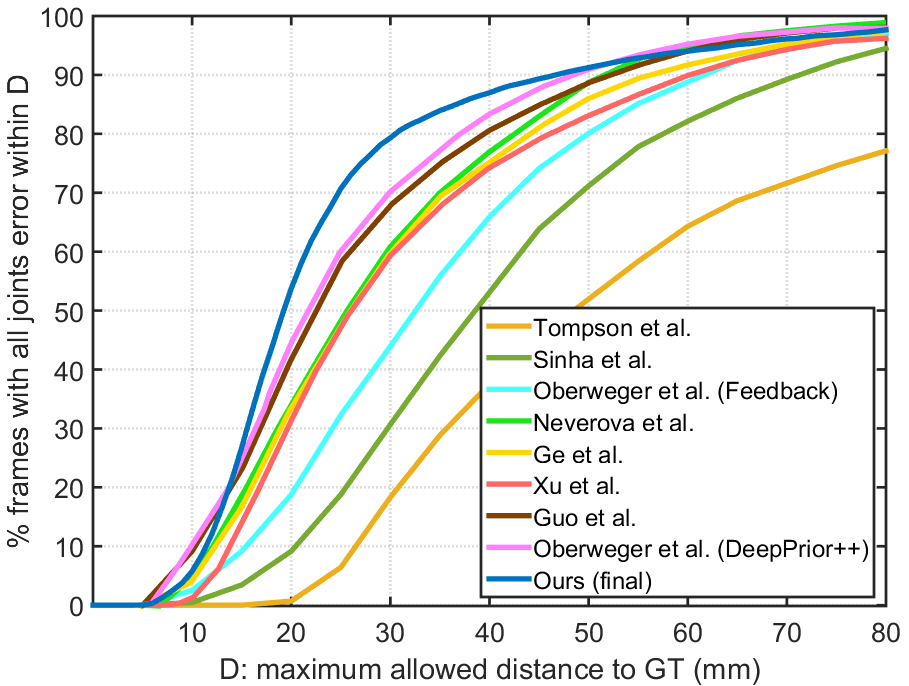}
  \label{fig:nyu-max}}
  \end{subfigure}\\
  \begin{subfigure}[]
  {\includegraphics[width=0.45\linewidth,height=4.8cm]{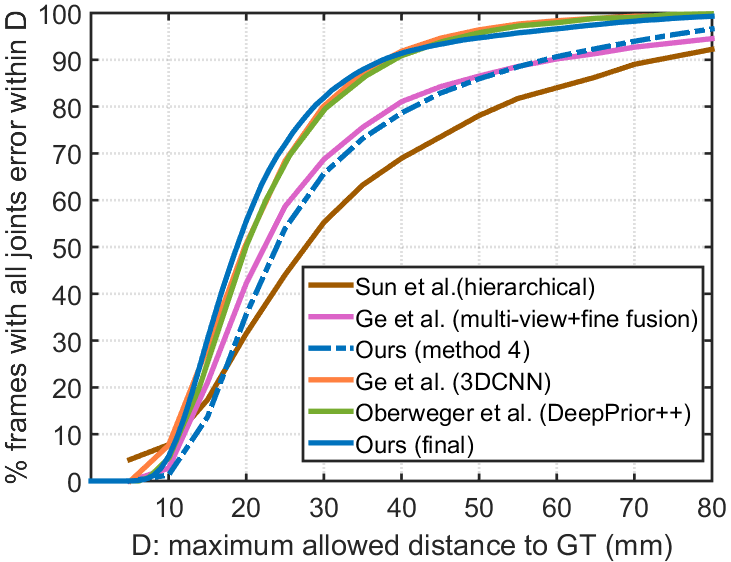}
  \label{fig:msra}}
  \end{subfigure}
  \begin{subfigure}[]
  {\includegraphics[width=0.45\linewidth,height=4.8cm]{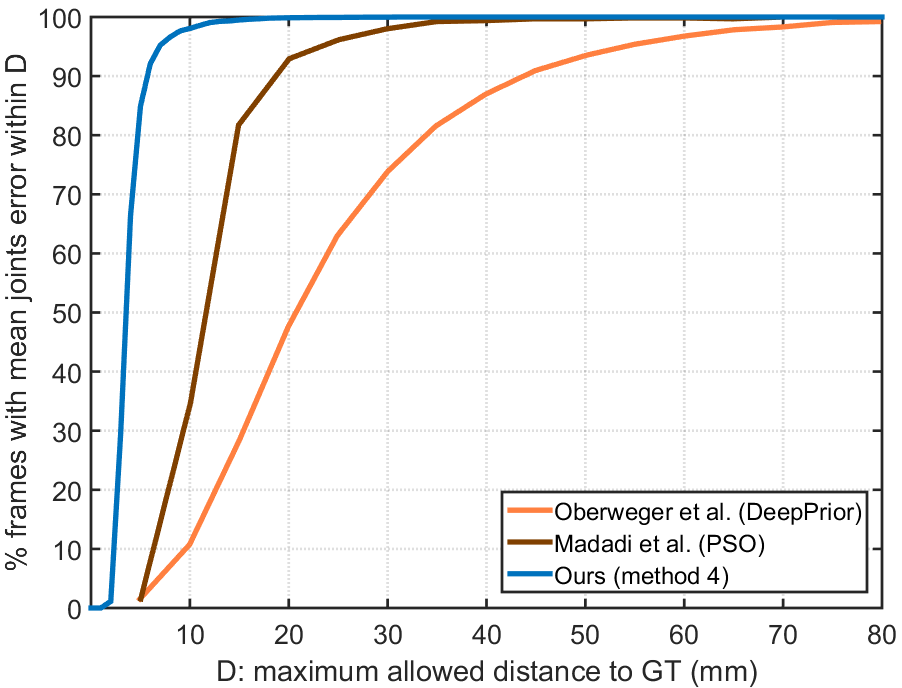}
  \label{fig:synthetic-mean}}
  \end{subfigure}
  \caption{State-of-the-art comparison. a) and b) Mean and maximum error success rate on NYU dataset. c) Maximum error success rate on MSRA dataset. d) Mean error success rate on SyntheticHand dataset.}
\end{figure*}

\subsection{Comparison with state of the art}
We report method the performance of our final model comparing to state-of-the-art data-driven approaches like \cite{tompson2014}, \cite{Oberweger2015iccv}, \cite{neverova2015hand}, \cite{sinha2016deephand}, \cite{ge20173d}, \cite{guo2017region}, \cite{xu2017lie} and \cite{oberweger2017deepprior++} on NYU dataset. On MSRA dataset we compare to \cite{Sun2015}, \cite{ge2016}, \cite{wan2017crossing}, \cite{ge20173d} and \cite{oberweger2017deepprior++}. Finally, we compare to \cite{Oberweger2015} and \cite{madadiFG2017} on SyntheticHand dataset. 

\textbf{NYU dataset.} Mentioned works in the comparison use 14 joints (as proposed in \cite{tompson2014}) to compare on NYU dataset. For a fair comparison on this dataset we take 11 joints most similar to \cite{tompson2014} out of our 20 used joints. We show maximum error success rate results in Fig.~\ref{fig:nyu-max}. As one can see, we outperform state-of-the-art results. However, \cite{guo2017region} and \cite{oberweger2017deepprior++} performs slightly better for error thresholds lower than 13mm. We also illustrate the average error success rate in Fig. \ref{fig:nyu-mean}. This shows our method is performing well in average for a majority of frames, i.e. less than 10mm error for 60\% of the test set. We compare to state-of-the-art regarding overall mean error in table \ref{tab:avg-err-nyu}. All these results show a significant improvement using data augmentation.

\begin{table}
\centering
\begin{tabular}{|l|c|}
\hline
Method & Average 3D \\
 & error (mm) \\ \hline
Oberweger et al. \cite{Oberweger2015} (DeepPrior) & 19.8 \\
Oberweger et al. \cite{Oberweger2015iccv} (Feedback) & 16.2 \\
Neverova et al. \cite{neverova2015hand} & 14.9 \\
Guo et al. \cite{guo2017region} (Ren) & 13.4 \\
Oberweger et al. \cite{oberweger2017deepprior++} (DeepPrior++) & 12.3 \\ \hline
Ours (\textit{4:3+viewpoint+fusion}) & 15.6 \\
Ours (final) & \bf{11.0} \\ \hline 
\end{tabular}
\caption{Average 3D error on NYU dataset.}
\label{tab:avg-err-nyu}
\end{table}

\begin{table}
\centering
\begin{tabular}{|l|c|}
\hline
Method & Average 3D \\
 & error (mm) \\ \hline
Sun et al. \cite{Sun2015} & 15.2 \\
Wan et al. \cite{wan2017crossing} (CrossingNet) & 12.2 \\
Oberweger et al. \cite{oberweger2017deepprior++} (DeepPrior++) & \bf{9.5} \\ \hline
Ours (\textit{4:3+viewpoint+fusion}) & 12.9 \\
Ours (final) & 9.7 \\ \hline 
\end{tabular}
\caption{Average 3D error on MSRA dataset.}
\label{tab:avg-err-msra}
\end{table}

\begin{figure*}[!t]
  \centering
  \begin{subfigure}[]
  {\includegraphics[width=0.75\linewidth]{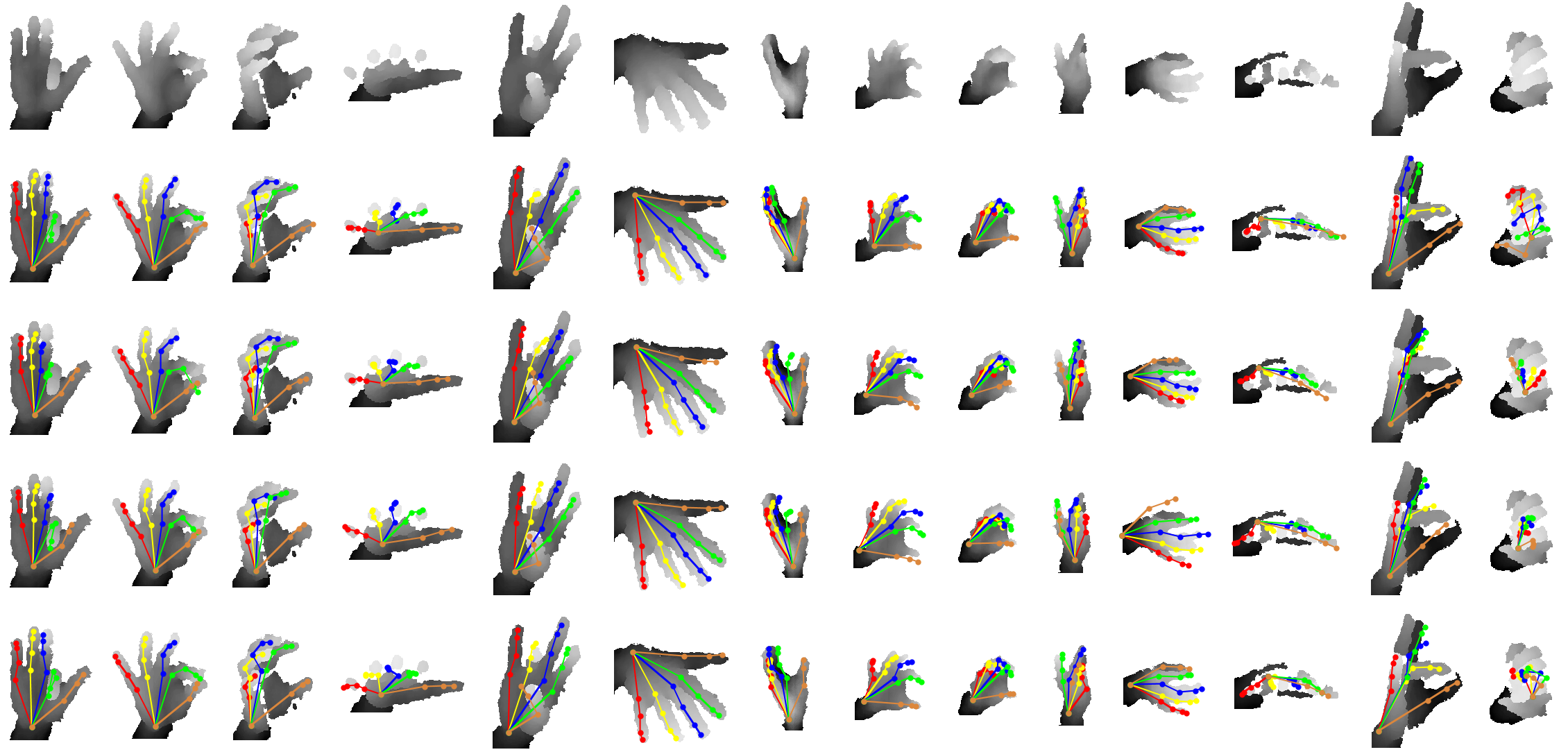}
  \label{fig:qualitative-nyu}}
  \end{subfigure}\\
  \begin{subfigure}[]
  {\includegraphics[width=0.75\linewidth]{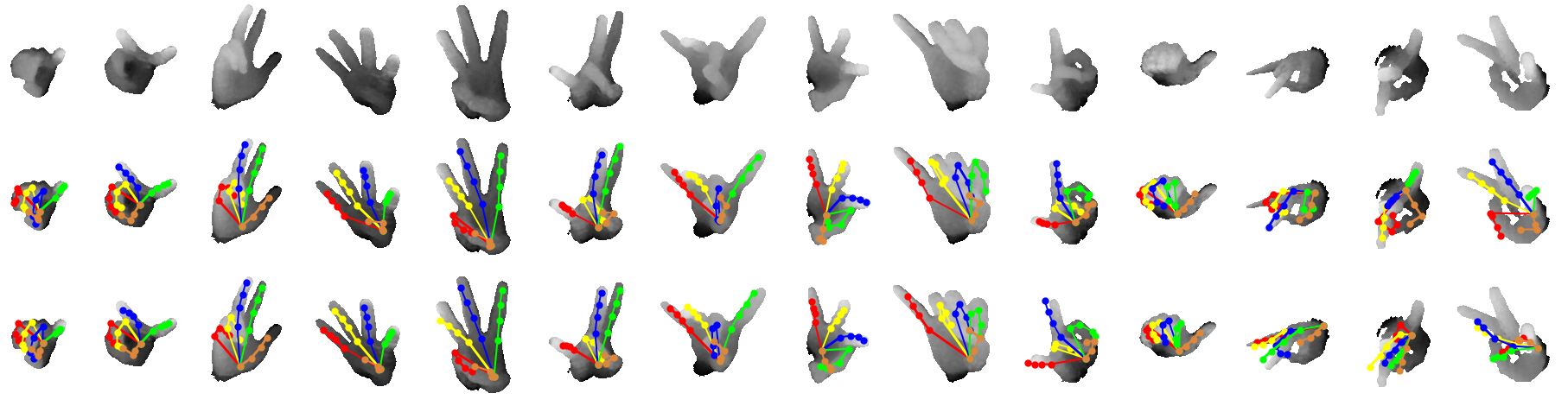}
  \label{fig:qualitative-msra}}
  \end{subfigure}\\
  \begin{subfigure}[]
  {\includegraphics[width=0.75\linewidth]{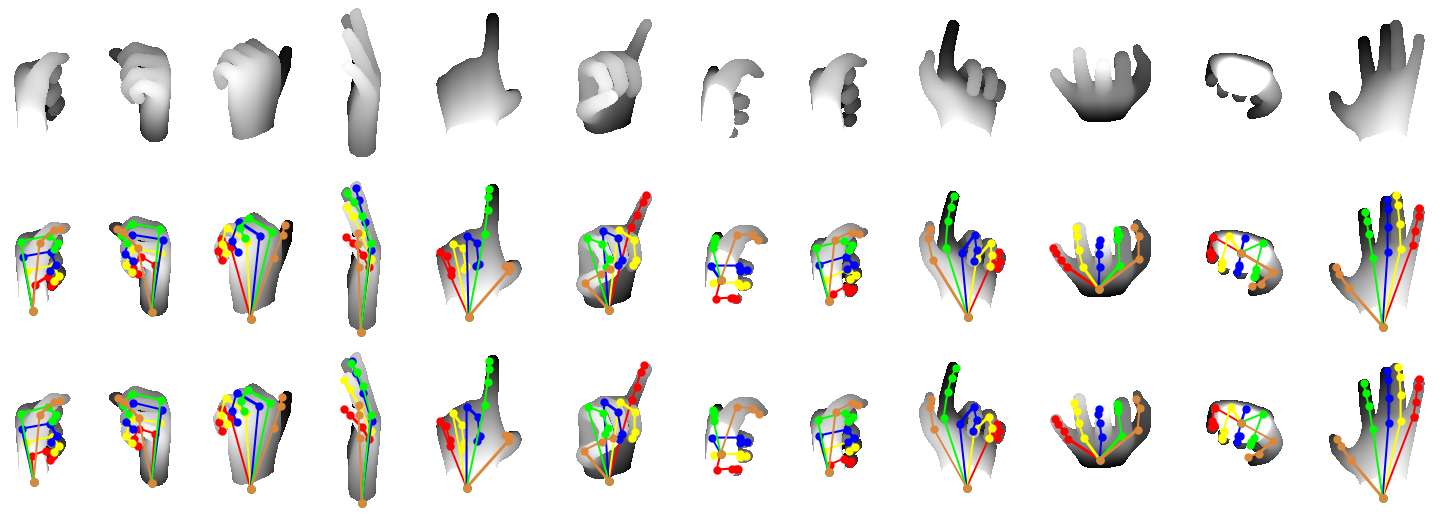}
  \label{fig:qualitative-synthetic}}
  \end{subfigure}
  \caption{Qualitative results. a) NYU dataset. Rows from top to bottom: depth image, ground truth, single channel network, method \textit{4:3+viewpoint+fusion} and final solution. Last three columns show maximum error higher than 50mm. b) MSRA dataset. Rows from top to bottom: depth image, ground truth and final solution. Last three columns show maximum error higher than 50mm. Ground truth annotations show inaccurate ground truth for this dataset especially for thumb. 12th column has inaccurate ground truth for little finger. c) SyntheticHand dataset. Rows from top to bottom: depth image, ground truth and final solution. See text for details of the methods.}
\end{figure*}

\textbf{MSRA dataset.} We applied introduced non-rigid hand augmentation same as NYU dataset. However, we observed a divergence during training. A possible reason could be the accuracy of ground truth annotations in MSRA dataset. Therefore, we applied standard augmentation techniques such as random scaling (in range $[0.9,1.05]$) and rotation (in range $[-90,90]$ degrees). We show the maximum error success rate results in Fig.~\ref{fig:msra}. As it can be seen, our method slightly outperform methods in the comparison for the error threshold between 13mm and 40mm. Although Sun \etal \cite{Sun2015} has a higher number of good frames for errors lower than 11mm, it performs the worst for higher error rates. Without using data augmentation, our method (dashed blue line) performs slightly worse than \cite{ge2016}. Note that \cite{ge2016} uses a pre-alignment over samples given hand point cloud eigenvectors which can be assumed as a kind of augmentation. We also show average error in table \ref{tab:avg-err-msra}. In average, our method with standard augmentation performs slightly similar to \cite{oberweger2017deepprior++} in this dataset. Note that \cite{oberweger2017deepprior++} uses random translation in the augmentation as well. We show some qualitative results in Fig. \ref{fig:qualitative-msra}. As one can see, ground truth annotations are not accurate in some cases, more specifically for thumb.

\textbf{SyntheticHand dataset.} We use original training set without augmentation to train our model on this dataset. Our model converges in 7 epochs. Mean error success rate is shown in Fig. \ref{fig:synthetic-mean}. As it can be seen, our method performs quite well on this dataset even for complex poses and viewpoints. Some qualitative results are shown in Fig. \ref{fig:qualitative-synthetic}. The overall average error on this dataset is 3.94mm.

\section{Conclusions}
\label{sec:conclusions}

We proposed a novel hierarchical tree-like structured CNN for recovering hand poses in depth maps. In this structure, branches are trained to become specialized in predefined subsets of the hand joints. We fused a network based on learned local features to model higher order dependencies among joints. The network is trained end-to-end. By including a new loss function incorporating appearance and physical constraints about doable hand motion and deformation, we found our network helps to increase the precision of the final hand pose estimations for quite challenging datasets. In particular, we found fusion network can help to better localize joints for easier hand configurations while it behaves similar to a local solution for more complex cases. We improved palm joints by applying a viewpoint regressor, and by fusing its learned features into the global pose. Finally, we introduced a non-rigid hand augmentation technique to deform original hands in terms of shape and pose helping to generalize better to unseen data. As a result we significantly improved estimations on original NYU dataset by 4.6mm in average. As future work, we will consider more complex data augmentation techniques to cope with noise in the depth image. Realistic data can be combined with synthetic data as well. In this sense, we will work on filling gaps realistically when more complex pose deformations are applied in the augmentation.

\section*{Acknowledgements}
This work has been partially supported by the Spanish projects TIN2015-65464-R and TIN2016-74946-P (MINECO/FEDER, UE) and CERCA Programme / Generalitat de Catalunya. We gratefully acknowledge the support of NVIDIA Corporation with the donation of the Titan Xp GPU used for this research.

{\small
\bibliographystyle{ieee}
\bibliography{egbib}
}

\end{document}